\documentclass[letterpaper]{article} % DO NOT CHANGE THIS
\usepackage{aaai25}  % DO NOT CHANGE THIS
\usepackage{times}  % DO NOT CHANGE THIS
\usepackage{helvet}  % DO NOT CHANGE THIS
\usepackage{courier}  % DO NOT CHANGE THIS
\usepackage[hyphens]{url}  % DO NOT CHANGE THIS
\usepackage{graphicx} % DO NOT CHANGE THIS
\usepackage{natbib}  % DO NOT CHANGE THIS AND DO NOT ADD ANY OPTIONS TO IT
\usepackage{caption} % DO NOT CHANGE THIS AND DO NOT ADD ANY OPTIONS TO IT
\usepackage{algorithm}
\usepackage{algorithmic}

\usepackage{newfloat}
\usepackage{listings}

\usepackage{booktabs}

\DeclareCaptionStyle{ruled}{labelfont=normalfont,labelsep=colon,strut=off} % DO NOT CHANGE THIS
\floatstyle{ruled}
\newfloat{listing}{tb}{lst}{}
\floatname{listing}{Listing}
\title{Simulation-based Scenario Generation for Robust Hybrid AI for Autonomy}
\author{
    Hambisa Keno, Nicholas J. Pioch\textsuperscript{\rm 1}\\
    Christopher Guagliano, Timothy H. Chung\textsuperscript{\rm2}
}
\affiliations{
    \textsuperscript{\rm 1}Systems \& Technology Research, LLC, 600 W. Cummings Pk, Woburn, MA 01801 USA\\
    \textsuperscript{\rm 2}Microsoft Corporation, 1 Microsoft Way, Redmond, CA 98052\\

    Hambisa.Keno@str.us, Nick.Pioch@str.us, cguagliano@microsoft.com, timothychung@microsoft.com
}

\usepackage{bibentry}
\begin{document}

\maketitle

\begin{abstract}
Application of Unmanned Aerial Vehicles (UAVs) in search and rescue, emergency management, and law enforcement has gained traction with the advent of low-cost platforms and sensor payloads. The emergence of hybrid neural and symbolic AI approaches for complex reasoning is expected to further push the boundaries of these applications with decreasing levels of human intervention. However, current UAV simulation environments lack semantic context suited to this hybrid approach. To address this gap, HAMERITT (Hybrid Ai Mission Environment for RapId Training and Testing) provides a simulation-based autonomy software framework that supports the training, testing and assurance of neuro-symbolic algorithms for autonomous maneuver and perception reasoning. HAMERITT includes scenario generation capabilities that offer mission-relevant contextual symbolic information in addition to raw sensor data. Scenarios include symbolic descriptions for entities of interest and their relations to scene elements, as well as spatial-temporal constraints in the form of time-bounded areas of interest with prior probabilities and restricted zones within those areas. HAMERITT also features support for training distinct algorithm threads for maneuver vs. perception within an end-to-end mission run. Future work includes improving scenario realism and scaling symbolic context generation through automated workflow.
\end{abstract}


\section{Introduction}

\begin{figure}[t]
\centering
\includegraphics[width=0.5\textwidth]{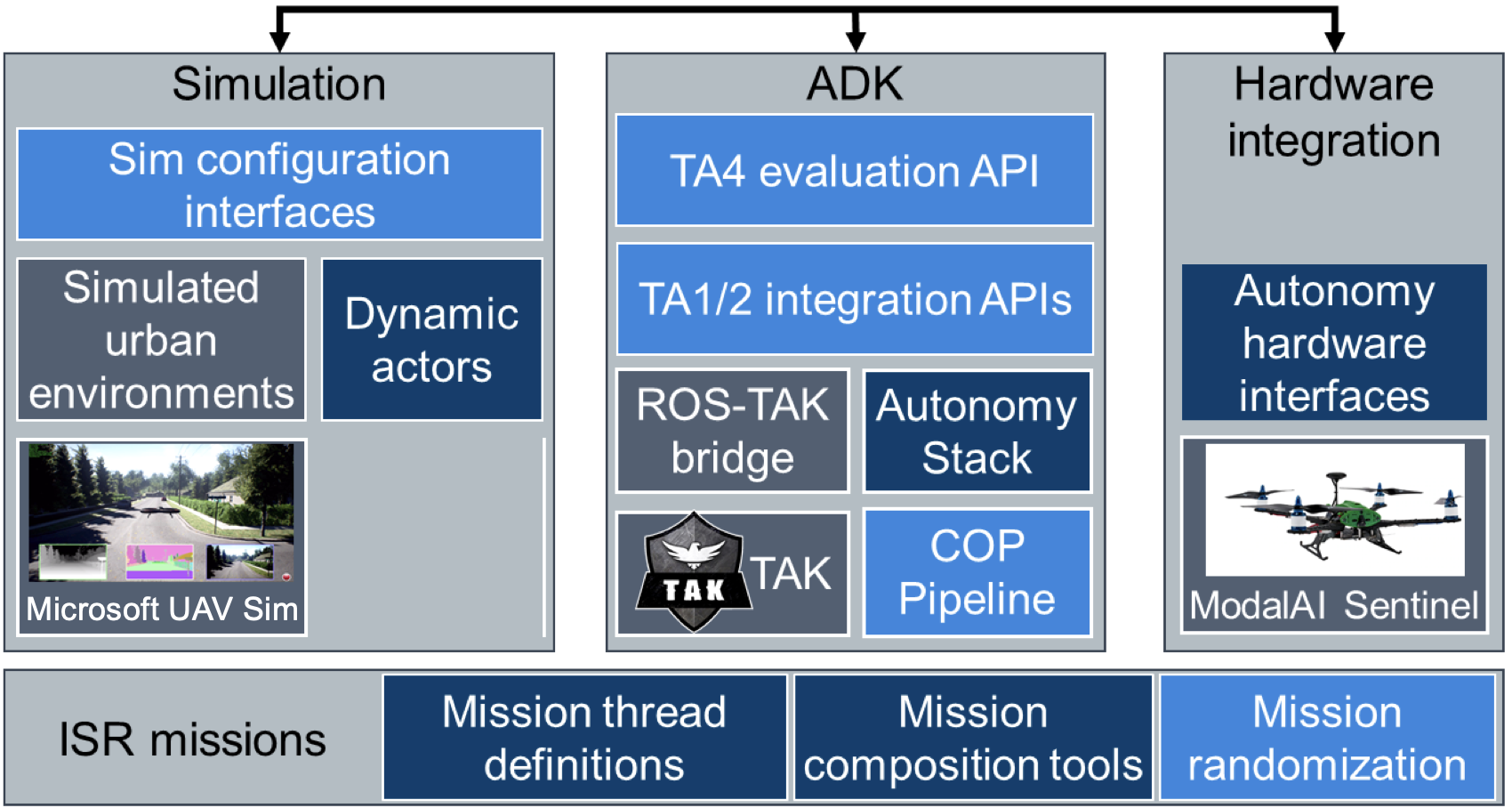} % Reduce the figure size so that it is slightly narrower than the column.
\caption{The HAMERITT architecture includes 1) Scenario \& Data Generation with tools for definition, composition, and randomization, 2) a high-fidelity Microsoft UAV simulator, and 3) an Algorithm Development Kit (ADK) with a baseline autonomy stack and COP-building pipeline with APIs tailored to ANSR performer roles. }
\label{fig:architecture}
\end{figure}

Under DARPA’s Assured Neuro-Symbolic Reasoning and Learning (ANSR) program \cite{c:24}, STR and Microsoft Corporation have developed a simulation-based autonomy framework entitled HAMERITT (Hybrid Ai Mission Environment for RapId Training and Testing). Figure \ref{fig:architecture} depicts HAMERITT’s high-level architecture, which is composed of four modules. This paper focuses primarily on the Scenario \& Data Generation module, which includes tools to define mission guidance, configure the underlying simulation for perception and maneuver mission threads, and support scenario randomization. HAMERITT uses an interactive Microsoft UAV simulator based on the former AirSim open source platform \cite{airsim2017, airsim2022} to provide enhanced environmental realism and simulation complexity. The target platform is a small UAV such as ModalAI’s Sentinel (Figure \ref{fig:architecture}, right) \cite{sent2024}. An Algorithm Development Kit (ADK) implements Application Programming Interfaces (APIs) in the Robot Operation System 2 (ROS 2) to expose  simulated data suited to symbolic reasoning, such as mission objectives, spatiotemporal constraints, and other contextual knowledge. The ADK API also exposes sensor data suited to neural machine learning, such as UAV camera feeds, depth map, and odometry (Figure \ref{fig:architecture}, center). The ADK includes support for training and testing ANSR performer algorithms (TA1 = algorithms, TA2 = assurance) and metrics generation in concert with the government evaluation team’s test harness and adversarial AI red teaming (Figure \ref{fig:architecture}, left). A pipeline for building a Common Operational Picture (COP) will support integrating maneuver and perception models into an end-to-end system, including exposure of the COP to end users via a ROS 2 to tak.gov bridge. Collectively, these capabilities enable HAMERITT clients to train against a vast variety of multi-modal scenarios to develop robust maneuver and perception algorithms for UAV urban search missions. In Phase 2 of ANSR, HAMERITT will integrate the performer algorithms into an overall ANSR system that provides a highly accurate Common Operational Picture for increasingly challenging scenarios, and transition the technology via integration with tak.gov, an open-source situational awareness framework widely used across military and civilian communities of interest.

The ADK was used by performer teams to train and test hybrid AI models for an interim Phase 1 evaluation. This HAMERITT release included diverse scenarios for single UAV search missions for entities of interest in urban environments, leveraging the Microsoft simulator to present challenging environmental conditions and perturbations (e.g., snow, fog, foliage, wind, time of day). The ADK provided support for two complementary mission threads (perception and maneuver) and included additional AI tools such as an automated path planner and fine-grained navigation between client-provided maneuver waypoints via Nav2 (see https://docs.nav2.org/).

\newpage
The following sections describe HAMERITT’s approach to scenario and data generation, simulation, and data for the interim evaluation. The last two sections discuss future research under ANSR TA3 and concluding remarks. 

\section{Scenario and Data Generation}\label{sec:scenarioAndDataGen}
\subsection{Scenario Specification}
HAMERITT’s unique contribution as a simulation-based autonomy environment is the multi-modal scenario data it provides, driven by mission objectives. The mission objective dictates the symbolic mission description data, the configuration of target entities including their trajectories, UAV sensor payloads, UAV trajectories and the environment in the simulation. Accordingly, upon requesting a random scenario from the HAMERITT system, two JSON-formatted files are generated: the mission description JSON, and the simulation configuration JSON. The mission description JSON is a structured representation of the mission objective, spatial-temporal constraints, prior beliefs in target entity locations, and scene description around target entities. The simulation configuration JSON specifies the starting state and trajectories for all mobile entities in the scene, the environmental conditions, the UAV payload configuration, and perturbation settings used to promote robustness of trained models.
\iffalse
\begin{figure}[H]
\centering
\includegraphics[width=0.5\textwidth]{figures/structured-scenario-rep.png} % Reduce the figure size so that it is slightly narrower than the column.
\caption{HAMERITT's structured scenario representation enables principled approaches for scenario randomization, and symbolic information augmentation via standardized operators}
\label{fig:structuredScenarioRep}
\end{figure}
\fi

\begin{figure}[H]
\centering
\includegraphics[width=0.5\textwidth]{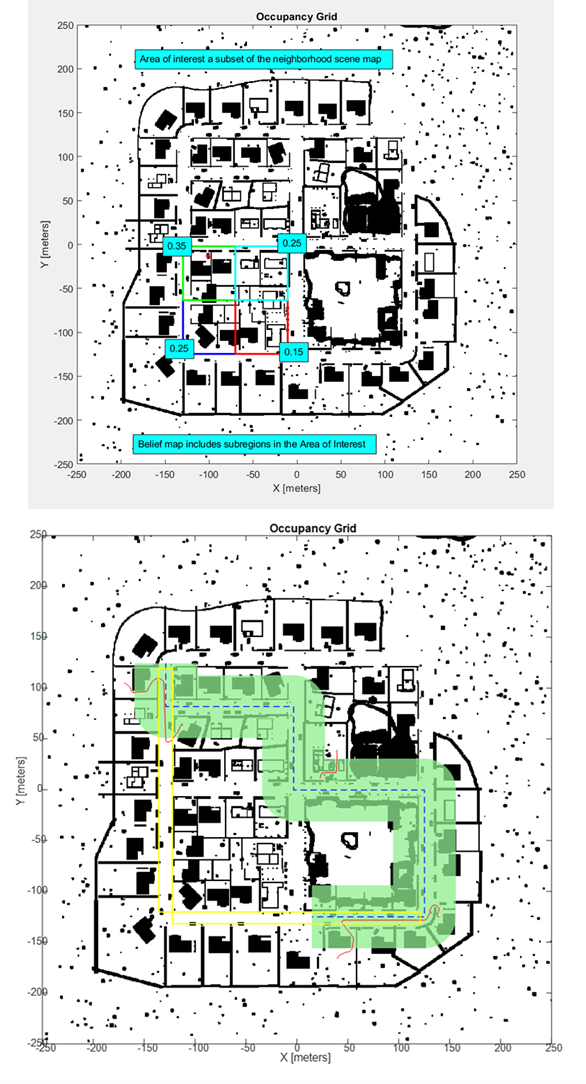} % Reduce the figure size so that it is slightly narrower than the column.
\caption{HAMERITT supports Area Search (top) and Route Search (bottom) scenario classes, with contextual symbolic information for spatial-temporal constraints and prior entity location belief maps.}
\label{fig:combinedOccGrid}
\end{figure}

HAMERITT currently supports two classes of scenarios. The \textit{Area Search Scenario} has the objective of locating a target of interest within one or more regions on the map designated as areas of interest (AOIs). In Figure \ref{fig:combinedOccGrid} (top), the target of interest is in the south-west quadrant on the map. The scenario description JSON represents this as the AOI for the search mission. The polygonal representation of the AOI provides symbolic cues for perception AI client solutions by providing a layer of disambiguation between entities. For example, to disambiguate two cars of interest with identical colors and models, the symbolic guidance could specify that one of them is outside the AOI. Similarly, maneuver AI client solutions can use the AOI as guidance for formulating search patterns. The scenario description file for area search further extends the set of symbolic context information by breaking up the AOI into sub-quadrants. The probability that a sub-quadrant contains the target entity is included as part of the mission description JSON as a prior on target entity location. Optionally, an AOI can be bound to a time window, providing a temporal disambiguation dimension.

\newpage
HAMERITT also generates \textit{Route Search} scenarios. The route search scenario has the objective of locating targets of interest on or along the periphery of a route of interest. The route of interest is represented as a sequence of line segments, while the periphery is represented by a search band as shown in Figure \ref{fig:combinedOccGrid} (bottom).

\iffalse
\begin{figure}[t]
\centering
\includegraphics[width=0.3\textwidth]{figures/route_search.png} % Reduce the figure size so that it is slightly narrower than the column.
\caption{Route search scenario. [EXPAND TO ACTION CAPTION].}
\label{fig:routeSearch}
\end{figure}
\fi
Both the area search and the route search scenarios are designed for dense urban environment to present challenges for both maneuver and perception algorithms. Occlusions from trees and buildings provide limited field-of-view for cameras on the UAV. The limited detection range for the cameras compounds this challenge by requiring UAVs to execute low-altitude flights in a crowded environment. Similarly, partial occlusions and images at arbitrary orientations due to UAV rotational maneuvers add to the perception challenge. The perception reasoner is also expected to fuse neural information, such as images containing a car, with symbolic information on which region that car is expected to be found. They also may need to recognize contextual objects in the scene around the car against the guidance provided as pre-mission symbolic context. For example, they may be asked to find a red sedan that has a garage to its right side. 

HAMERITT includes vignettes for   \textit{Moving Target Pursuit}. These vignettes are aimed at increasing the complexity of the challenge problems. The maneuver challenge for these vignettes involves achieving camera pose to place the dynamic target in view continuously. The perception challenge involves reasoning about the target in the presence of confusers and reacquiring the target when temporarily out of view. 
\begin{figure}[t]
\centering
\includegraphics[width=0.5\textwidth]{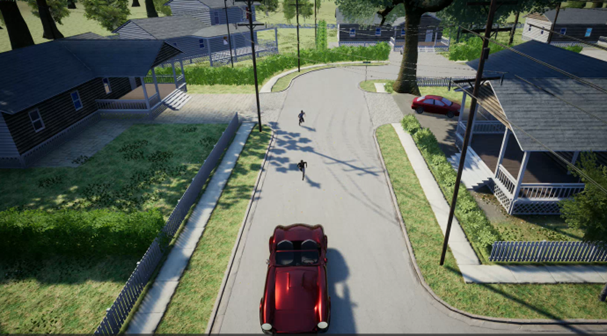} % Reduce the figure size so that it is slightly narrower than the column.
\caption{HAMERITT includes vignettes to support reasoning about complex events. The vignette showing two pedestrians chasing  the red car provide an instance of anomalous scene. The subsequent trajectory followed by the red car enables a pursuit challenge for a UAV.}
\label{fig:kozAndCarTraj}
\end{figure}

\subsection{Contextual symbolic information}
As discussed above, the scenario description file provides a set of contextual symbolic information that are intended to augment the neural data stream from sensors. This information consists of two main types: constraints or scene descriptions.

HAMERITT uses three types of constraints in specifying a scenario. The \textit{Area of Interest (AOI)} is a soft-spatial constraint. In the area search context, it specifies the region of the map where the target of interest is expected to be found. A solution for searching this area of interest can fly the UAV outside of the AOI itself without violating this constraint. Optionally, the AOI constraint can have a no-later-than (NLT) and a no-earlier-than (NET) temporal constraint specifying the valid period for the spatial AOI constraint.

The \textit{Keep Out Zone (KOZ)} constraint is a hard spatial constraint. A UAV flight path that intersects one or more keep out zone polygons violates this constraint. Like the AOI constraint, we can apply optional temporal bounds to the KOZ constraints. HAMERITT uses the KOZs as levers for introducing both maneuver and perception challenges. For example, the KOZs can restrict UAV avenues of approach to target entity that provide occlusion free vantage points. HAMERITT also uses a combination of time bound KOZs and known target entity trajectories to restrict UAV access to portions of the entity trajectory. For example, to prevent a UAV detecting a target car while the car is on a particular street, the polygon bounding the street, and the projected interval for the car on the street can be mapped into KOZ and time bound on the KOZ respectively. Figure \ref{fig:combinedOccGrid} (bottom) depicts this use case where we deny the UAV access to two streets (yellow boxes) going from North to South and West to East at the precise time interval where two cars of interest will traverse a portion of their trajectories (red curves) along those streets.

\textit{Area Priors} are sub-regions on the map with probabilities for containing the target entity. We can loosely interpret these priors as soft spatial constraints. That is, they can be considered as a high-resolution variation of the AOI where they represent a more fine-grained belief on where the target of interest is expected to be found.  Figure \ref{fig:combinedOccGrid} (top) shows an example of these Area Priors where the sub quadrant that contains the entity of interest has the highest prior. This symbolic context information can inform explicit search strategy or search policy generated from learning-based methods, for example a more frequent revisit of the sub-quadrant with the highest prior until target detection.
\iffalse
\begin{figure}[t]
\centering
\includegraphics[width=0.5\textwidth]{figures/koz-car-traj.png} % Reduce the figure size so that it is slightly narrower than the column.
\caption{Keep out zones for street access denial.}
\label{fig:kozAndCarTraj}
\end{figure}
\fi
In addition to constraints, HAMERITT provides scene descriptions as a second class of contextual symbolic information. The purpose of these scene descriptions is to enhance the performance of a perception reasoner in the presence of ambiguities. For example, a search scenario can ask for locating a blue Chevy Impala in a one square mile area covering multiple city blocks. A UAV flying over this area can detect multiple instances of blue Chevy Impalas. However, if the perception reasoner is provided additional descriptions about the scene around the particular blue Chevy Impala, it can augment the raw sensor data with this scene description to disambiguate the target car from similar looking cars within the perimeter of the city blocks of interest. 

\begin{figure}[t]
\centering
\includegraphics[width=0.5\textwidth]{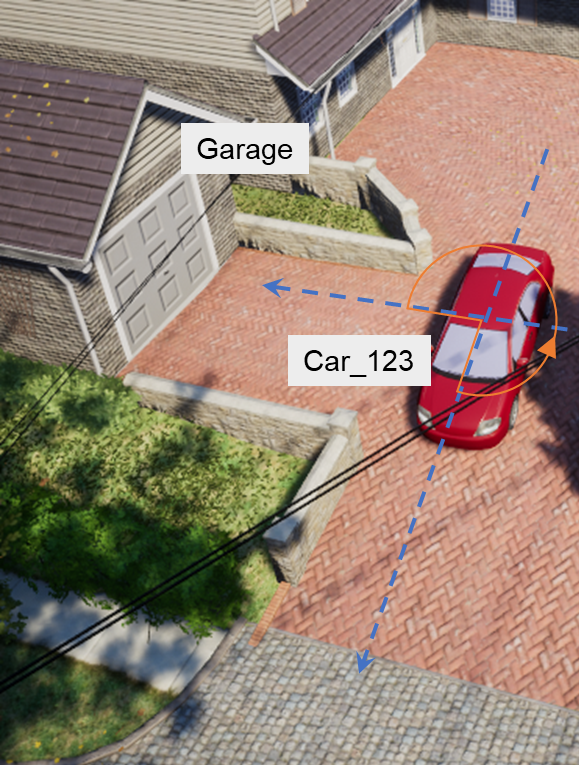} % Reduce the figure size so that it is slightly narrower than the column.
\caption{A "RIGHT\_OF" spatial relationship between target car (Car\_123) and a garage.}
\label{fig:symbExample}
\end{figure}

HAMERITT uses a simple list of symbolic operators to represent the relationship between a target entity and its surrounding scene environment. A relationship has the following format. [Related entity] [relationship operator] [Target entity] [(Optional) Related entity attributes]. For example, the symbolic relationship shown in Figure \ref{fig:symbExample} is represented as [Garage][RIGHT\_OF][Car\_123][Garage\_color: White, Garage\_number\_of\_doors: 1]. 

\begin{table}
  \caption{Symbolic relationship definition}
  \label{tbl:symbolicRelations}
  \centering
  \begin{tabular}{p{1.2in}p{1.5in}}
    \toprule
     Operator     & Description \\
    \midrule
    (NOT)\_NEXT\_TO  & Distance between Target Entity and Related Entity less than x meters with no Line-of-site obstruction in between   \\
         (NOT)\_ON\_TOP\_OF & Minimum Z value for bounding box for Target Entity approximately equals max Z value for the bounding box for the Related Entity     \\
         ORTHOGONAL\_TO      & Line-of-sight to Related Entity with respect to body frame for Target Entity approximately 90-degrees  \\
         IN\_FRONT\_OF & Line-of-sight to Related Entity with respect to body frame for Target Entity approximately 180-degrees \\
         RIGHT\_OF & Line-of-sight to Related Entity with respect to body frame for Target Entity approximately 270-degrees \\
         LEFT\_OF & Line-of-sight to Related Entity with respect to body frame for Target Entity approximately 90-degrees \\
         PART\_OF & The Target Entity is an element of the Related Entity. In this case, the Related is an abstract concept such as a group number. \\
         EVENTUALLY\_\$SO & Specifies the relationship defined using the spatial operators will be true in an indeterminate future. \\
    \bottomrule
  \end{tabular}
\end{table}

Table \ref{tbl:symbolicRelations} shows the list of symbolic operators HAMERITT has instantiated so far. These operators describe one of three types of relationships – spatial, membership, or spatial-temporal. The spatial relationship operator definition uses the Target Entity’s body frame as the reference with counterclockwise rotation with respect to the Target Entity yaw assumed positive. The “Eventually\_\$SO” operator combines temporal information with one of the spatial relationships. 

\section{Simulation Environment}\label{sec:simulationEnv}

Microsoft has developed an advanced UAV simulator platform for HAMERITT to meet the ANSR program’s mission-focused needs. The extended platform exposes a common interface and communication layer for robotics applications via the ROS 2 pub/sub framework, allowing rapid integration and coordination of performer algorithms for maneuver and perception. To support versatility of scenes and rapid scenario configuration, the Microsoft simulator leverages the widely used Unreal Engine for high-fidelity physics and graphics. The simulator supports simulation of complex UAS search tasks and environment conditions spanning normal and anomalous traffic and pedestrian behaviors in urban scenes. 

\begin{figure}[t]
\centering
\includegraphics[width=0.5\textwidth]{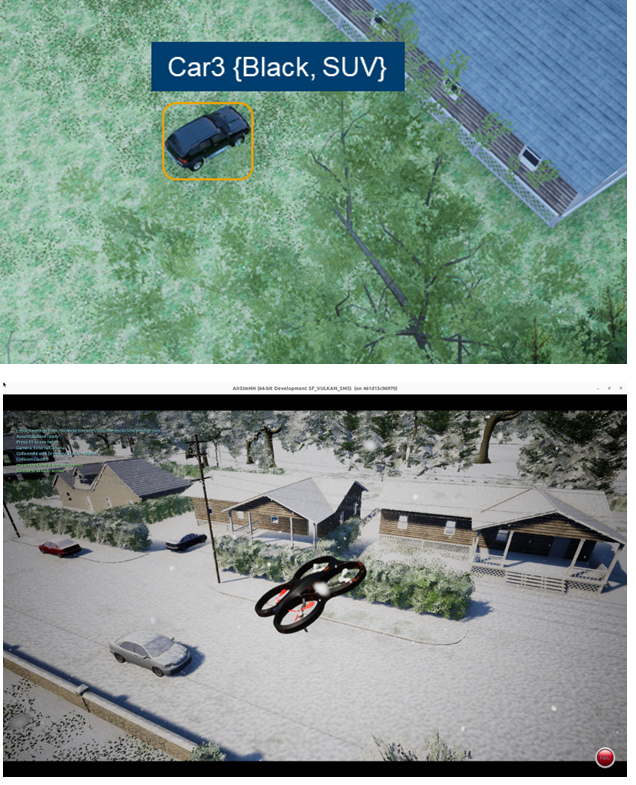} % Reduce the figure size so that it is slightly narrower than the column.
\caption{HAMERITT uses a realistic Microsoft UAV simulator to dynamically spawn entities of interest (top) and simulate realistic environmental variations including challenging weather and visibility conditions (bottom).}
\label{fig:dynamicSpawn}
\end{figure}

Figure \ref{fig:dynamicSpawn} (top) shows an example scene generated by the Microsoft simulator within HAMERITT. One of the target entities of interest (Car 3) is parked in the backyard, with semantic attributes for color and vehicle make provided as symbolic context. Figure \ref{fig:dynamicSpawn} (bottom) shows a particularly challenging weather condition simulating light snow, which can obscure the true color of a target vehicle, presenting a robustness challenge for perception modeling. Additional variations to foster model robustness include wind, camera noise, time of day variations, and occlusion.

\section{Dataset for evaluation}\label{sec:dataSetForEval}

HAMERITT generated more than 9,000 scenario variations for area search and route search in a suburban neighborhood environment developed under previous research \cite{offset2023, ccast2021, offset-c2}. The resulting dataset has variability along the UAV location, UAV flight altitude, UAV flight path, weather conditions for snow, rain, dust fog, wind, foliage, and camera noise settings. HAMERITT also included the route search scenario variation with dynamic target entities in the scene. An offline path planner generates a set of obstacle avoiding routes for the target car entities. A similar approach is used to generate a safe-maneuver path for the UAV to enable training for perception-only reasoning. The safe-maneuver paths for the perception-only thread are generated such that they guarantee a look at the entities of interest. These offline path planners use the Rapidly exploring Random Tree (RRT) algorithm \cite{rrt2011} which is well suited for generating path variability. HAMERITT also integrated the Nav2 stack for waypoint following to allow real-time global path planning based on sparse waypoint inputs. To support maneuver-only training, HAMERITT provides perfect perception feedback via reporting of ground truth location information on an entity of interest. HAMERITT triggers the perfect perfection report when the maneuver solution was successful in positioning the UAV such that the target entity is in the frame for one or more cameras onboard the UAV.
\section{Future extensions}\label{sec:futureExt}

Future extensions to our work include the following. First, we will extend HAMERITT APIs to enable integration of hybrid AI models for the partial mission threads into an end-to-end system. Second, we plan on increasing the complexity of scenarios from search only to include search, follow, track, and avoid tasks, while migrating from a suburban neighborhood to a denser city environment. Third, we plan to automate the scenario generation and randomization scheme and increase the variety of training conditions using UC Berkeley's SCENIC framework \cite{fremont2019scenic}. Fourth, we will expose HAMERITT's Common Operating picture (COP) to end-users via a ROS 2 interface for the tak.gov platform to support transition to DoD and law enforcement communities. Finally, we plan on increasing the fidelity of the UAV dynamics through PX4 integration \cite{px2015} to enhance robustness of hybrid AI maneuver models in advance of live field tests for sim2real transfer evaluation.  

\section{Conclusion}\label{sec:conclusion}
In conclusion, HAMERITT provides an interactive simulation-based data generation framework to foster development of robust hybrid neuro-symbolic AI models for perception and maneuver. It provides scenario configuration tools for specifying semantic mission objectives, contextual knowledge, and environmental conditions. The HAMERITT ADK exposes UAV data feeds and ground truth via a ROS 2 interface, including pixel-level camera and depth imagery, with semantic labels and target entity bounding boxes for AI model training. A Microsoft UAV simulation environment is used to create realistic scenes and behaviors embodying challenging urban search tasks, with configurable scripting of target entity locations and behaviors, and a range of environmental variations including weather, time of day, visibility, and more. For the ANSR program Evaluation 1, HAMERITT delivered more than 9000 scenario variations for Area Search and Route Search in a typical neighborhood environment. Future work includes expansion to more complex missions, increased automation for scenario generation, integration of perception and maneuver models into an end-to-end system, and transfer to a physical UAV platform for live field tests.

\section{Acknowledgments}
Distribution Statement A (Approved for Public Release, Distribution Unlimited).

\noindent The views, opinions, and/or findings expressed are those of the author(s) and should not be interpreted as representing the official views or policies of the Department of Defense or the U.S. Government.

\bibliography{aaai25}

\begin{thebibliography}{10}
\providecommand{\natexlab}[1]{#1}

\bibitem[{Chung and Daniel(2023)}]{offset2023}
Chung, T.; and Daniel, R. 2023.
\newblock {DARPA OFFSET: A Vision for Advanced Swarm Systems through Agile Technology Development and Experimentation}.
\newblock \emph{Field Robotics}, 3: 97--124.

\bibitem[{Clark et~al.(2021)Clark, Usbeck, Diller, and Schantz}]{ccast2021}
Clark, S.; Usbeck, S.; Diller, D.; and Schantz, R. 2021.
\newblock {CCAST: A Framework and practical deployment of heterogeneous unmanned system swarms}.
\newblock \emph{GetMobile: Mobile Computing and Communications}, 24(4): 17--26.

\bibitem[{{DARPA}(2024)}]{c:24}
{DARPA}. 2024.
\newblock Assured Neuro Symbolic Learning and Reasoning (ANSR).
\newblock \url{https://www.darpa.mil/program/assured-neuro-symbolic-learning-and-reasoning}.
\newblock Accessed: 2024-07-15.

\bibitem[{Fremont et~al.(2019)Fremont, Dreossi, Ghosh, Yue, Sangiovanni-Vincentelli, and Seshia}]{fremont2019scenic}
Fremont, D.~J.; Dreossi, T.; Ghosh, S.; Yue, X.; Sangiovanni-Vincentelli, A.~L.; and Seshia, S.~A. 2019.
\newblock Scenic: a language for scenario specification and scene generation.
\newblock In \emph{Proceedings of the 40th ACM SIGPLAN conference on programming language design and implementation}, 63--78.

\bibitem[{Lorenz, Honegger, and Pollefeys(2015)}]{px2015}
Lorenz, M.; Honegger, D.; and Pollefeys, M. 2015.
\newblock PX4: A node-based multithreaded open source robotics framework for deeply embedded platforms.
\newblock In \emph{2015 IEEE international conference on robotics and automation (ICRA)}, 6235--6240.

\bibitem[{ModalAI(2024)}]{sent2024}
ModalAI. 2024.
\newblock \emph{https://www.modalai.com/products/sentinel-development-drone?variant=40446488805427}.

\bibitem[{Sertac et~al.(2011)Sertac, Walter, Perez, Frazzoli, and Teller}]{rrt2011}
Sertac, K.; Walter, M.; Perez, A.; Frazzoli, E.; and Teller, S. 2011.
\newblock Anytime motion planning using the RRT.
\newblock In \emph{2011 IEEE international conference on robotics and automation}, 1478--1483.

\bibitem[{Shah et~al.(2017)Shah, Dey, Lovett, and Kapoor}]{airsim2017}
Shah, S.; Dey, D.; Lovett, C.; and Kapoor, A. 2017.
\newblock Airsim: High-fidelity visual and physical simulation for autonomous vehicles.
\newblock In \emph{Field and Service Robotics: Results of the 11th International Conference}, 621--635.

\bibitem[{Wang et~al.(2022)Wang, Nir, Vemprala, Kapoor, Ofek, McDuff, and Gonzalez-Franco}]{airsim2022}
Wang, C.; Nir, O.; Vemprala, S.; Kapoor, A.; Ofek, E.; McDuff, D.; and Gonzalez-Franco, M. 2022.
\newblock CityLifeSim: A High-Fidelity Pedestrian and Vehicle Simulation with Complex Behaviors.
\newblock In \emph{2022 IEEE 2nd International Conference on Intelligent Reality (ICIR)}, 11--16.

\bibitem[{Williamson et~al.(2023)Williamson, Taranta, Moolenaar, and LaViola}]{offset-c2}
Williamson, B.; Taranta, E.; Moolenaar, Y.; and LaViola, J. 2023.
\newblock {Command and Control of a Large Scale Swarm Using Natural Human Interfaces}.
\newblock \emph{Field Robotics}, 3: 301--322.

\end{thebibliography}
\end{document}